\documentclass[10pt,twocolumn,letterpaper]{article}

\usepackage[accsupp]{axessibility}

\usepackage{wacv}
\usepackage{times}
\usepackage{epsfig}
\usepackage{graphicx}
\usepackage{amsmath}
\usepackage{amssymb}

\usepackage{times}
\usepackage{soul}
\usepackage{url}

\usepackage{amsmath}
\usepackage{amsthm}
\usepackage{booktabs}
\usepackage{algorithm}
\usepackage{algorithmic}
\usepackage{mathtools}
\usepackage{comment}
\usepackage{multirow}
\usepackage{amsfonts}
\usepackage{bigstrut}

\def\wacvPaperID{1389} 

\wacvfinalcopy

\ifwacvfinal
\usepackage[breaklinks=true,bookmarks=false]{hyperref}
\else
\usepackage[pagebackref=true,breaklinks=true,colorlinks,bookmarks=false]{hyperref}
\fi

\begin{document}

\title{Variational Stacked Local Attention Networks for Diverse Video Captioning}

\author{\centering Tonmoay Deb \and Akib Sadmanee \and Kishor Kumar Bhaumik \and Amin Ahsan Ali \and M Ashraful Amin \and A K M Mahbubur Rahman
\and {\small Artificial Intelligence and Cybernetics (AGenCy) Lab, Independent University, Bangladesh}\\

{\tt\small tonmoay.nsu@gmail.com,  \{1620274, 1621366, aminali, aminmdashraful, akmmrahman\}@iub.edu.bd}
}

\maketitle
\thispagestyle{empty}

\begin{abstract}
	While describing spatio-temporal events in natural language, video captioning models mostly rely on the encoder's latent visual representation. Recent progress on the encoder-decoder model attends encoder features mainly in linear interaction with the decoder. However, growing model complexity for visual data encourages more explicit feature interaction for fine-grained information, which is currently absent in the video captioning domain. Moreover, feature aggregations methods have been used to unveil richer visual representation, either by the concatenation or using a linear layer. Though feature sets for a video semantically overlap to some extent, these approaches result in objective mismatch and feature redundancy. In addition, diversity in captions is a fundamental component of expressing one event from several meaningful perspectives, currently missing in the temporal, i.e., video captioning domain. To this end, we propose Variational Stacked Local Attention Network (VSLAN), which exploits low-rank bilinear pooling for self-attentive feature interaction and stacking multiple video feature streams in a discount fashion. Each feature stack's learned attributes contribute to our proposed diversity encoding module, followed by the decoding query stage to facilitate end-to-end diverse and natural captions without any explicit supervision on attributes. We evaluate VSLAN on MSVD and MSR-VTT datasets in terms of syntax and diversity. The CIDEr score of VSLAN outperforms current off-the-shelf methods by $7.8\%$ on MSVD and $4.5\%$ on MSR-VTT, respectively. On the same datasets, VSLAN achieves competitive results in caption diversity metrics. 
\end{abstract}

\section{Introduction}

\begin{figure*}[t]
	\centering
	\includegraphics[width=0.94\textwidth]{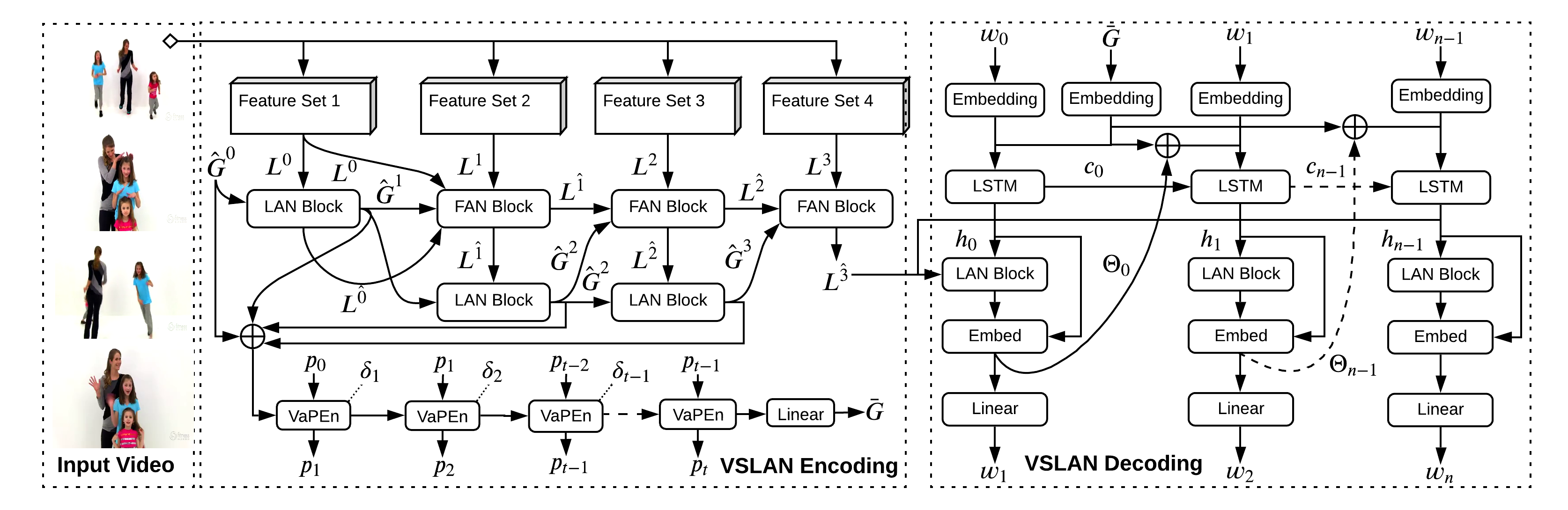}

	\caption{Overview of Variational Stacked Local Attention Network (VSLAN) comprising of Local Attention Network (LAN), Feature Aggregation Network (FAN), and Variational POS Encoder (VaPEn). The final Local feature set from FAN is utilized to query caption words at the decoding stage, done by another dedicated LAN block. VaPEn is built on top of the output of each FAN block. The final output from VaPEn's final node accounts for global POS feature, $\overline G$, further added with ${{\Theta _{n}}}$ for influencing the diversity on decoder.}
	\label{fig:slan_model}
\end{figure*}

With the trailing success of Recurrent Neural Networks (RNN), sequence-to-sequence (seq2seq) \cite{venugopalan2014translating} was introduced, which incorporates Long Short Term Memory (LSTM), a variant of RNN to generate captions from encoded visual features of Convolutional Neural Network (CNN). However, addressing the drawback of global features (by seq2seq) on missing important temporal context, attention mechanism \cite{yao2015describing} was proposed in spatio-temporal domain to improve the decoder during step-by-step word prediction. Later several improvements on attention mechanism \cite{pu2018adaptive,chen2019motion} have evolved for processing visual features to capture both spatial and temporal relation effectively.

Most existing works exploit attentive features from the encoder stage as hidden input to the decoder. Although it is a widely used norm, it is limited to perform in a single feature stream. Because the attention distribution is uni-modal and weights all feature streams in an identical fashion. To tackle this issue, several studies from visual recognition \cite{kong2017low} and question-answering \cite{kim2018bilinear} studied capturing richer feature semantics by facilitating better interaction. Bilinear Pooling \cite{gao2016compact} have been widely studied to be efficiently computed regardless of the number of interaction order with a cost of expensive computation. Recently, \cite{kim2016hadamard} proposed linearly mapped low-rank bilinear pooling using the Hadamard product, which solved this issue. Following that, this concept is adopted by several domains \cite{xlan}. We similarly incorporate bilinear pooling as a building block of our local and global attention module. On top of that, we propose a novel self-attention architecture to attend specific feature sets directly from training loss. To our best knowledge, the concept of exploiting temporal structure, e.g., for video captioning, for generating fine-grained feature set with bilinear pooling, is yet unexplored.

Another research track focuses on visual feature sophistication with an intuition that richer features can lead to subtle clues for guiding the decoder. A straightforward approach is to capture features from different architectures and unify them into a fixed dimension. The networks can be frame-level 2D CNN and clip-level 3D CNN features. We argue that word prediction models can better differentiate between the candidate vocabularies when features from multiple origins are fed into the decoder. Also, the aggregation approach of the works above fuse features at a single stage, either by concatenation or with a linear layer. Yet, hierarchical dependence between the streams was ignored. In this scenario, a feature stream can hardly identify attributes already addressed by the prior streams, resulting in redundancy. Given that feature, compression \cite{wang2018reconstruction}, and model distillation \cite{mmsibnet} are active research domains, unifying redundant features from multiple streams is worthwhile to explore. This work proposed a module that attends to specific feature sets from respective streams in a forward pass and learns to discount the redundant information end-to-end without explicit supervision.

Amid notable progress in video captioning, the main focus was on generating deterministic captions, i.e., single caption output from a video. Although one caption is expected to represent overall temporal dynamics, it is rarely true from the real-world perspective. Moreover, generating multiple-yet-relevant captions can semantically diversify the overall visual understandings from a video, which may also be useful in video query and retrieval. However, extracting probabilistic, i.e., diverse captions from videos, is also an unexplored problem to the best of our knowledge. Although some progress has been made for static images \cite{jyoti}, the main challenge for videos remains in attending to a temporal window and encoding the respective features into a meaningful representation. To tackle this issue, we include diverse Parts of Speech (POS) generation to influence the final caption structure. Because POS information holds the basic pattern of grammar composition to control the final sentence structure \cite{jyoti, wang2019controllable} and can be easily evaluated with existing metrics, e.g., BLEU \cite{bleupaper}. Thus, we propose an approach to generate diverse POS from the fine-grained feature streams. The final representation from POS further works as a global feature of the caption decoder that also influences sentence diversity.

In summary, we propose Variational Stacked Local Attention Network (VSLAN), comprising of Local Attention Network (LAN), Feature Aggregation Network (FAN) modules to exploit the potential of visual features from several off-the-shelf pre-trained models during captioning. The LAN block utilizes bilinear pooling to attend the related clips of a video during captioning, results in richer and highly attended information. While a new feature set arrives, the FAN block aggregates them so that it discounts previously learned information from preceding LAN blocks, which ensures compact visual features. The LAN-FAN blocks are propagated by aggregating specialized visual feature sets and leveraged into a decoder for captioning. We enforce diversity in sentences by proposing a Variational POS Encoder (VaPEn) using the fine-grained representation from each feature stream. In the decoding stage, the local features from the last encoder layer are used to query the next words, whereas the global features, i.e., stochastic VaPEn features, influence that query stream for diverse caption syntax. Figure \ref{fig:slan_model} illustrates the primary view of VSLAN. Therefore, our key contributions are:

\begin{itemize}
	\item We propose a novel, end-to-end video captioning architecture, VSLAN, which can attend to both local (in each feature stream) and global (in-between feature streams) spatio-temporal attributes without explicit supervision to capture richer semantics for decoding robust and diverse captions.
	\item We reformulate the existing feature aggregation methods by introducing a novel network to propagate the important information from the new feature streams in a discount fashion.
	\item We introduce diversity in video captioning by formulating and utilizing an end-to-end POS generator with temporal context, VaPEn, to influence decoder query. 

	\item We perform extensive analysis to demonstrate the efficacy of our proposed architecture. VSLAN outperforms state-of-the-art methods by a notable margin on two datasets and attains consistent diversity performance based on commonly used evaluation protocol.
\end{itemize}

\section{Related Works}

\noindent\textbf{Sequence Learning and Attention}

approaches use encoder-decoder to learn a shared representation of visual and linguistic features.  
Recently, \cite{mmdenseatt} proposed densely connected recurrent neural networks to facilitate attention and improve accuracy. The SibNet architecture by \cite{mmsibnet} builds on top of the encoder-decoder network, where two CNN branches are utilized to encode the videos. Another track of attention-based video captioning focuses on finding out salient region \cite{mmsalient} in each video frame using attention and simultaneously learn the spatio-temporal representation over time. \cite{mmhmm} did it more explicitly by storing attributes in a hierarchical memory and utilize while captioning. However, due to vanishing gradient issue by recurrent networks, these approaches fall short for long-tail videos. In our work, we overcome this limitation by designing the caption decoder utilizing key-query-value representation for each recurrent node and training a shareable LAN module.

\begin{figure}[t]
	\centering
	\includegraphics[width=0.76\columnwidth]{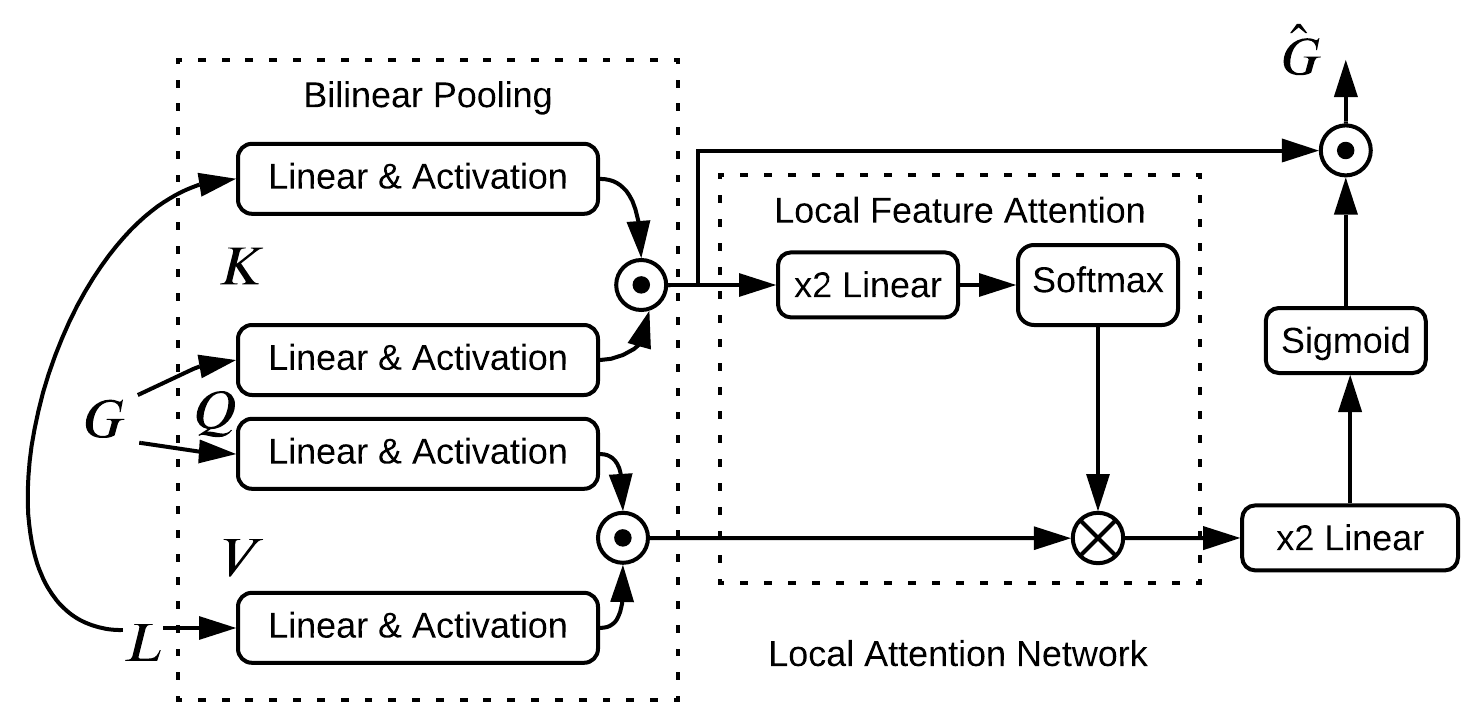}
	\caption{Overview of Local Attention Network (LAN).}
	\label{fig:lanblock}
\end{figure}

\noindent\textbf{Feature Aggregation:}
\cite{venugopalan2015sequence} used a hyperparameter to calculate the weighted sum of Optical Flow and RGB features at each step. \cite{chen2017video,hori2017attention} enhanced encoder representation by including multimodal, e.g., audio, category features along with the visual stream. \cite{wang2019controllable} designed cross-gating (CG) block to fuse RGB and Flow features toward captioning. Recently, 
to introduce feature invariance, \cite{mmtwice} proposed encoding videos twice to avoid irrelevant temporal information. We argue that this strategy will fall short on online video captioning. Our approach self-attends each feature stream and learns redundancy while among multiple streams, results in fine-grained aggregated features.

\noindent\textbf{Bilinear Pooling} in image captioning has been recently explored by \cite{xlan}. However, their inherent intuition and architecture are different from ours. For example, they applied squeeze-excitation \cite{squeeze} over their one of the key-query feature split. However, VSLAN consists of a straightforward element-wise multiplication between the key-query and the value's linearly pooled feature, extended with two linear layers and a sigmoid to connect the key-query representation. They utilize a learned decoder from the visual features for intra-modal (text and image) queries during decoding. In contrast, we use a dedicated LAN decoder, shared in the decoding stage by learning from scratch. 

\noindent\textbf{Diversity in Captioning} had initial success by deterministic Diverse Beam Search (DBS) \cite{dbs}. \cite{jyoti} recently incorporated a stochastic latent space learning like variational RNN (VRNN) \cite{vrnn} and applied it for predicting words from images. Although there have been no explicit experiments on videos, \cite{mmcontroldiverse} recently introduced an approach for generating captions from example sentences in addition to videos. Their model incorporates the syntax structure of the example sentence during caption generation. In this case, even controllable, it requires additional data to train and becomes unstable with unseen examples. Our diverse caption generation focuses more on inherent syntactic structure, i.e., POS generation, which attends to specific visual features and influences the decoder to generate diverse captions.

\section{Variational Stacked Local Attention Nets}

\subsection{Video Encoder}

Given $M = ({L^1},{L^2},...,{L^{M - 1}})$ a set of local feature set from a video, layer ${L^m} = \left( {l_1^m,l_2^m,...,l_N^m} \right)$, where $N$ is the video length and ${l_i^m}$ is extracted feature from $i$-th clip, the goal is to generate a description $S = \left( {{w_1},{w_2},...,{w_n}} \right)$ of ${n}$ length, by sampling each word, ${w}$, from the vocabulary $\vartheta $.

\begin{figure}[t]
	\centering
	\scalebox{0.85}{
		\includegraphics[width=0.8\columnwidth]{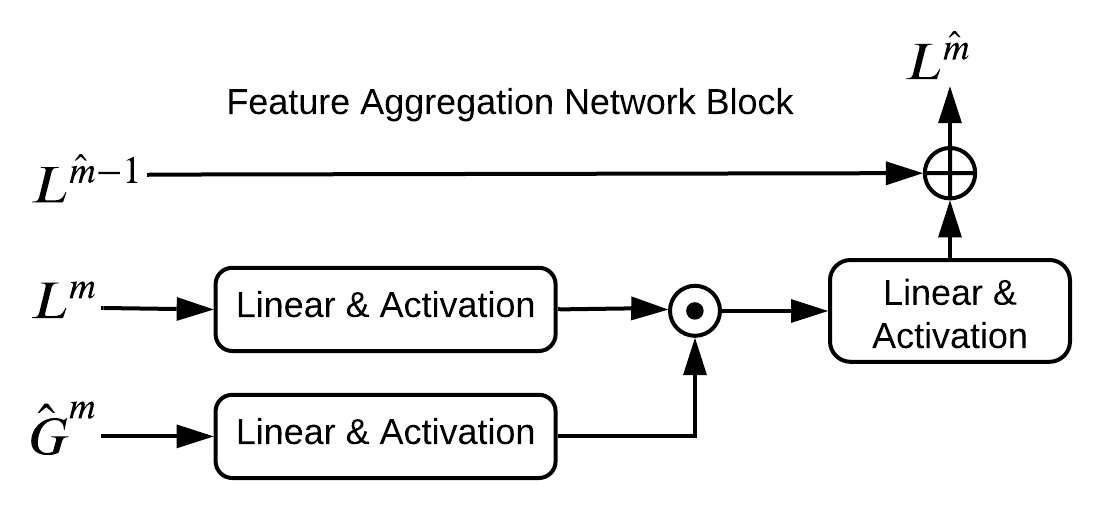}
	}
	\caption{Overview of Feature Aggregation Network (FAN).}
	\label{fig:fan_block}
\end{figure}

\noindent\textbf{Local Attention Network (LAN):} 

For Query $Q \in {\mathbb{R}^q}$, Key $K = \left( {{k_1},{k_2},...,{k_N}} \right)$, and Value $V = \left( {{v_1},{v_2},...,{v_N}} \right)$, where ${k_i} \in {\mathbb{R}^k}$ and ${v_i} \in {\mathbb{R}^v}$, we perform low-rank bilinear pooling on $Q$ and $K$ as:
\begin{equation}
	\label{eqn:1}
	\beta _i^K = \sigma \left( {W_Q^KQ} \right) \odot \sigma \left( {{W_K}{k_i}} \right)
\end{equation}

Here, $W_Q^K \in {\mathbb{R}^{z \times q}}$ and ${W_K} \in {\mathbb{R}^{z \times k}}$ are embedding metrics, projects $Q$ and $K$ into a unified dimension ${\mathbb{R}^z}$. $\sigma $ is the activation function. The projected $Q$ performs element-wise multiplication, $ \odot $, with each projected key ${k_i}$. Similarly, for Value $V$, the pooling
\begin{equation}
	\label{eqn:2}
	\beta _i^V = \sigma \left( {W_Q^VQ} \right) \odot \sigma \left( {{W_V}{v_i}} \right)
\end{equation}

Here, $W_Q^V \in {\mathbb{R}^{z \times q}}$ and ${W_V} \in {\mathbb{R}^{z \times v}}$ projects $Q$ and $K$ into a unified dimension ${\mathbb{R}^z}$. ${\beta ^K} = (\beta _1^K,\beta _2^K,...,\beta _N^K)$ and ${\beta ^V} = (\beta _1^V,\beta _2^V,...,\beta _N^V)$ are bilinear Key, Value attention distribution for network layer $m$, holding ${2^{{m^{th}}}}$ order feature interaction properties.

To capture semantic information, ${\beta ^K}$ and  ${\beta ^V}$ are propagated with two network streams to generate clip-wise, local attention, and global attention distribution, respectively. The local information $\beta _i^{s'}$ is captured as:

\begin{equation}
	\label{eqn:3}
	\beta _i^{s'} = \sigma \left( {W_{{\beta ^V}}^s\sigma \left( {W_{{\beta ^V}}^k\beta _i^K} \right)} \right)
\end{equation}

Here, $W_{{\beta ^V}}^k \in {\mathbb{R}^{z' \times z}}$ and $W_{{\beta ^V}}^s \in {\mathbb{R}^{1 \times z'}}$. The fist linear layer projects the local feature into a richer representation dimension $z'$, where the second lowers the dimension to $1$ for probabilistic local feature ${\beta ^s} = softmax\left( {{\beta ^{s'}}} \right)$

Later, a weighted sum is performed on bilinear pooled value ${\beta ^V}$ with the distribution ${\beta ^s}$ as ${\beta ^l} = \sum\limits_{i = 1}^N {\beta _i^s\beta _i^V} $. The attended local features represent key clips that are relevant to the captioning query $Q$. ${\beta ^l} \in {\mathbb{R}^{1 \times z}}$ is further passed through two linear layers, followed by a sigmoid to generate fine-grained global information ${\beta ^g} = \sigma \left( {W_{{\beta ^l}}^2\left( {W_{{\beta ^l}}^1{\beta ^l}} \right)} \right)$ 

Again, $W_{{\beta ^l}}^1 \in {\mathbb{R}^{x \times z}}$ and $W_{{\beta ^l}}^2 \in {\mathbb{R}^{z \times x}}$ are the projection metrics where $x < z$. The lower dimension of $x$ squeezes global information into a latent representation and reconstructs the previous dimension accordingly. $\sigma$ is sigmoid activation function. The concept of ${{W_{{\beta ^l}}^1{\beta ^l}}}$ is similar to feature compression technique, concerning robust representation at the decoding stage. ${\beta ^g} \in {\mathbb{R}^{1 \times z}}$ is further combined with key-query pairs, ${\beta _i^K}$, to compute stabilized global feature ${\hat G}$ with the network as $
\hat G = \sum\limits_{i = 1}^N {\beta _i^l\beta _i^K}
$

${\hat G}$ is the enhanced representation using self-attention, which incorporates local information with explicit feature interaction compared to the regular self-attention that utilizes only first-order properties as mentioned in Figure \ref{fig:lanblock}. We exploit this pooling strategy over multiple layers using feature aggregation, described below.

\noindent\textbf{Feature Aggregation Network (FAN):} 

Each FAN block uses low-rank bilinear pooling to capture novel information from the new feature stream and integrates with the current.

For pre-trained CNN features ${L^{\hat m - 1}}$, the global information is captured by ${{\hat G}^m}$ using LAN. When a new CNN feature set ${L^m}$ appears, a low-rank biliear pooling operation is being performed between ${L^m}$ and ${{\hat G}^{m}}$ as
\begin{equation}		
	\beta _i^L = \sigma \left( {W_G^m{{\hat G}^m}} \right) \odot \sigma \left( {W_L^mL_i^m} \right)
\end{equation}

Here, for layer $m$, $W_G^m \in {\mathbb{R}^{y \times z}}$ and $W_L^m \in {\mathbb{R}^{y \times v }}$ projects ${{{\hat G}^m}}$ and ${{L^m}}$ into a fixed dimension $y$ to perform $ \odot $ operation. $\beta _i^L$ captures the relevant information from ${L^m}$ by exploiting low-rank bilinear pooling using previous global feature ${{{\hat G}^m}}$. $\beta _i^L$ is further passed through a linear layer with activation for $\beta _i^{L'} = \sigma \left( {W_{L'}^m\beta _i^L} \right)$. Here, $W_{L'}^m \in {\mathbb{R}^{z \times y}}$ projects features into $z$ dimension. The final representation ${L^{\hat m}}$ from the FAN block is comprised of $LayerNorm$ over pairwise sum of projected feature $\beta _i^{L'}$ and previous attentive local features ${L^{\hat m - 1}}$

${L^{\hat m}} \in {\mathbb{R}^{N \times z}}$ is a bridge between the preceding and current local information, respectively, further passed to LAN block for richer, more explicit attention distribution, resulting distilled feature representation. For the initial FAN block, we consider ${L^{\hat m - 1}} = {L^m} = {L^0}$.

As discussed earlier, out of $M$ feature sets, FAN discounts redundant information present in the prior feature propagation. During the training stage, the network distributes non-linear attentive weights to the representation that are more relevant to the caption.  
Following the recommendation from \cite{aafaq2019spatio,hou2020joint}, we exploit object-wise and temporal convolutional networks from the uniformly sampled frames and clips to represent subject and object properties, respectively. We set the visual features in ${L^m}$ as mentioned in Figure \ref{fig:fan_block}, where LAN and FAN learn weights end-to-end from the caption loss without explicit guidance (fundamental mechanism of self-attention).

\noindent\textbf{Variational POS Encoder (VaPEn):} 
We generate stochastic POS sequences from the global visual features $\tilde G  = \left[ {{{\hat G}^0};{{\hat G}^1};...;{{\hat G}^{M - 1}}} \right] \in {\mathbb{R}^{q=(M - 1)*z}}$. Here, $\left[ {.;.} \right]$ denotes concatenation operation. $\tilde G$ is set at the intial hidden state to the encoder LSTM. We incorporate VRNN \cite{vrnn}, which learns the latent distributions $\delta_{t}$ over time steps $t$ of a recurrent network. During the generation stage, at first ${\delta _t} \sim {\mathcal N}({\mu _{0 ,t}},{\sigma _{0 ,t}})$ distribution is conditioned on the prior LSTM state variable $s_{t-1}$ and decodes POS $p_t|{\delta _t} \sim {\mathcal N}({\mu _{\delta ,t}},{\sigma _{\delta ,t}})$ as output using the distribution at that time step. Modifying the original Variational Autoencoder, the recurrent generator prior distribution ${r_\theta }({p_{ \le T}}|{\delta _{ \le T}}) = \prod\limits_{t = 1}^T {{r_\theta }({p_t}|{\delta _{ \le t}},{p_{ < t}})} {r_\theta }({\delta _t}|{p_{ < t}},{\delta _{ < t}})$. During the inference, $\delta _{t}$ is updated from the generated POS $p_t$ by following posterior distribution ${q_\varphi }({\delta _{ \le t}},{p_{ \le t}}) = \prod\limits_{t = 1}^T {{q_\varphi }({\delta _t}|{p_{ \le t}},{\delta _{ < t}})} $. Here $\theta$ and $\varphi$ are the recurrent network, i.e., LSTM parameters, updated end-to-end. The overall objective function comprises reaching variational lower bound based on prior and encoder according to the below objective:

\begin{equation*}
    \begin{split}
        {\mathbb{E}_{{q_\varphi}(\delta _{\leq T}|p_{\leq T})} \Bigg[
        {\sum\limits_{t=1}^T}(-KL({q_\varphi }({\delta _t}|{\delta _{ \leq t}},{p_{ \leq t}})
         ||{r_\theta }({\delta _t}|{\delta _{ < t}},{p_{ < t}}))}\\{ + \log {r_\theta }({p_t}|{\delta _{ \leq t},{p_{ < t}}}}
        \Bigg]
    \end{split}
\end{equation*}


Here $KL({q_\varphi }||{r_\theta })$ is Kullback–Leibler divergence between two distributions, $q_\varphi$ and $r_\theta$, measures nonsymmetric difference, aimed to maximize w.r.t the parameters. We refer readers to \cite{vrnn} for more details. The last hidden state, $s_{t}$ is further propagated to the decoder as and shared across the decoding steps as stated in the next section.
\subsection{Caption Decoder}

The fundamental goal of a caption decoder is to generate a sequence of words conditioned on the encoder features and preceding words. The decoder is comprised of basic $LSTM$ network with a decoder $LAN$ block, shared among all $LSTM$ nodes. With total $M$ number of feature sets, $m = {(M - 1)^{th}}$ layer information, ${L^{\hat m}}$ is passed to the decoding stage along with the mean projected pooled global information $\overline G  = {W_gs_{t}}$. Here ${W_g} \in {\mathbb{R}^{z \times q }}$, projects the final POS features into $z$ dimension. Particularly, for a timestep $t$, input to the decoder $LSTM$ is the embedding of word ${w_t}$, concatenated with element-wise sum of global feature $\overline G $ and bilinear-pooled representation ${\Theta _{t - 1}}$ is:
\begin{equation}
	{h_t},{c_t} = LSTM\left( {[{W_E}{w_t};{\Theta _{t - 1}} \oplus \overline G ],{h_{t - 1}},{c_{t - 1}}} \right)
\end{equation}

Here, ${h_t}$ and ${c_t}$ are the hidden output and memory state from $LSTM$, ${{W_E}}$ is the word-embedding matrix, and $ \oplus $ is the summation operation. ${{\Theta _{t}}}$ is calculated by inserting ${h_t}$ as Query $Q$ to the $LAN$ block with the Local Features ${L^{\hat m}}$ as ${\Theta _t} = {W_A}\left[ {LAN({h_t},{L^{\hat m}});{h_t}} \right]$

By exploiting $LAN$, ${\Theta _t}$ is the enhanced representation of $LSTM$ output, where it bridges the encoder information with the hidden states of $LSTM$ to bring the local visual information as keys and values attended by ${2^{nd}}$ order interaction. After linear projection using ${W_A}$, ${\Theta _t}$ is passed using an embedding layer of vocabulary dimension, followed by a softmax function to predict the next word as:
\begin{equation}
	{w_{t + 1}} = {p_\theta}(\left. {{w_{t + 1}}} \right|{w_{1:t}},\theta ) = softmax\left( {{W_\upsilon }\sigma \left( {{\Theta _t}} \right)} \right)
	\label{word_loss}
\end{equation}

The probabilistic word generation based on given parameters $\theta $ continues until a STOP token is received or generation reaches maximum caption length. Figure \ref{fig:slan_model} depicts overall captioning architecture.

\noindent\textbf{Entailment-based Caption Scoring:}
Textual entailment predicts the probability of a proposition to be true provided another proposition is true. Recent papers use CIDEr score as the reward function \cite{wang2019controllable}. However, CIDEr uses undirected n-gram, assigning a high score even if there is a small-sized but critical error like negation, unrelated action, or object.
According to \cite{Pasunuru2017ReinforcedVC}, textual entailment ensures that the hypothesis gets a high reward only if it is logically inferred from the premise. Therefore we use the entailment scores as reward $r(w^s)$ in equation \ref{eqn:loss} to only calculate the loss on standalone RL training. We use RoBERTa \cite{Liu2019RoBERTaAR} instead of the Decomposable Attention model (DA) \cite{Pasunuru2017ReinforcedVC} as DA assigns unreasonably high entailment scores to many non-matching sentence pairs. We present a comparative analysis of the CIDEr score and two entailment scores, DA \& RoBERTa in Table \ref{table:entailment}. RoBERTa captures the wrong object and action better than DA.  

\begin{table}[t]
	\centering
	\small

	\scalebox{0.62}{

		\begin{tabular}{|llcrr|}
			\hline
			\multicolumn{1}{|c}{\multirow{2}[4]{*}{\textbf{Ground Truth}}} & \multicolumn{1}{c}{\multirow{2}[4]{*}{\textbf{Generated Caption}}} & \multirow{2}[4]{*}{\textbf{CIDEr}} & \multicolumn{2}{c|}{\textbf{Entailment score}} \bigstrut\\
			\cline{4-5}      &       &       & \multicolumn{1}{c}{\textbf{DA}} & \multicolumn{1}{c|}{\textbf{RoBERTa}} \bigstrut\\
			\hline
			\hline
			a person is playing a game & a man is playing a baseball & \multirow{4}[2]{*}{0.24} & 0.999 & 0.003 \bigstrut[t]\\
			a young girl on stage & a boy is singing on stage &       & 1     & 0.006 \\
			a man narrates a pokemon game & a person is playing a video game &       & 0.971 & 0.753 \\
			a man is playing a video game & a man is talking about a river &       & 0.999 & 0.012 \bigstrut[b]\\
			\hline
		\end{tabular}

	}	
	\caption{Comparison of CIDEr and entailment scores.}
	\label{table:entailment}
\end{table}

\noindent\textbf{Shared Learning Strategy:} 

Given  $S = \left( {{w_1},{w_2},...,{w_n}} \right)$ as ground truth sequence for a clip, the goal is to learn model parameters $\theta $ and minimize Cross-Entropy (XE) loss:
\begin{equation}
	{L^{XE}}(\theta ) =  - \sum\limits_{t = 1}^n {p(\left. {{w_t}} \right|{w_{1:t - 1}},\theta )}
\end{equation}

However, according to recent studies \cite{mixer}, word-level cross-entropy training restricts sentence generation in a specific structure. 
For training VSLAN, we employ Self-critical Sequence Training \cite{scstpaper} to minimize the expected reward, sentence-pair similarity score discussed in the previous section. The loss function:
\begin{equation}
	\label{eqn:loss}
	{\nabla _\theta }{L^{RL}}(\theta ) \approx  - (r({w^s}) - b){\nabla _\theta }\log {p_\theta }({w^s})	
\end{equation}

Here, ${r({w^s})}$ is the reward for sampled sentence ${w^s}$ with a baseline $b$ for variance reduction.
 
Standalone RL-based training does not warrant caption reliability \cite{rlnotbetter}, and fluency. 
We experiment with shared learning strategy, which employs both cross-entropy and reinforcement learning algorithm to calculate a shared loss ${L^S}(\theta )$:
\begin{equation}
	\label{eqn:shared_learn}
	{L^S}(\theta ) = \eta {L^{XE}}(\theta ) + (1 - \eta ){L^{RL}}(\theta )
\end{equation}

Here, $\eta$ is an adjustable hyperparameter. We evaluate VSLAN  with ${L^{XE}}$, ${L^{RL}}$, and ${L^S}$ to demonstrate the performance benchmark of the loss functions, respectively.

\section{Experiments}

\subsection{Dataset Details and Preprocessing}
We evaluate VSLAN on MS Research Video Description (MSVD/ YouTube2Text) \cite{msvdpaper} and MSR-VTT \cite{msrvttpaper}. More details on these datasets and the preprocessing strategies can be found in the attached \textbf{supplementary material}.

\subsection{Implementation Details}

\textbf{Feature Extraction:}  For object-level features, we utilize Faster R-CNN \cite{ren2015faster}. For a clip ${l_i}$, 30 frames are uniformly sampled and sorted based on the number of predicted boxes. The frame with the highest number of boxes is used to extract the regional features that represents the objects of that clip. We also used an off-the-shelf 2D feature descriptor, VGG16 \cite{vggpaper}, to extract the mean vector from $5$ uniformly sampled frames from the clip. As the captions are mostly focused on the video activities, we leverage 3D CNNs i.e., C3D \cite{c3dpaper} and ResNext-101 \cite{resnext}. Both architectures are trained on 3D kernel to capture the temporal information for each video clip. For both networks, we use pre-trained models based on the Kinetics-400 dataset. We assume that, due to the inherent architectural difference, they produce distinguishable representations even trained on an identical dataset. The clip frames have been resized into $256 \times 256$ dimensions for network input.

For each clip, outputs of the pre-$softmax$ layers are used as feature of the 3D CNN models. We initially set ${L^0}$= ResNext-101, ${L^1}$= C3D, ${L^2}$= VGG16, and ${L^3}$= Faster R-CNN to update the network parameters from coarse action representation to fine-grained object information. Moreover, we alternate the order of feature sets by holdout or shuffle and discuss result changes in the \textbf{supplementary material}.

\begin{table}[t]
	\small
	\centering

	\scalebox{0.82}{

		\begin{tabular}{|c|c|c|c|c|c|}
			\hline
			\textbf{Model} & \textbf{Backbone} & \textbf{B@4} & \textbf{M} & \textbf{C} & \textbf{R} \bigstrut\\
			\hline
			\hline
			\multicolumn{6}{|c|}{\textit{Encoder-Decoder with Attention}} \bigstrut\\
			\hline
			DenseLSTM & C3D+V   & 50.4 & 32.9 & 72.6 & - \bigstrut[t]\\
			SAM & R & 54  & 35.3  & 87.4  & - \\
			VRE & R & 51.7  & 34.3  & 86.7  & 71.9 \\
			X-LAN & FR    & 52.1  & 31.1  & 79.9  & 72.2 \bigstrut[b]\\
			\hline
			\hline
			\multicolumn{6}{|c|}{\textit{Explicit Attribute Guidance	}} \bigstrut\\
			\hline
			GRU-EVE & IR+C3D+Y & 47.9  & 35    & 78.1  & 71.5 \bigstrut[t]\\
			HMM & C3D+IR & 52.9  & 33.8  & 74.5  & - \\
			C-R   & IR+R & \underline{57} & \underline{36.8} & \underline{96.8} & - \bigstrut[b]\\
			\hline
			\hline
			\multicolumn{6}{|c|}{\textit{Encoder-Decoder with Reconstruction}} \bigstrut\\
			\hline
			RecNet\textsubscript{local} & V+IR  & 52.3  & 34.1  & 80.3  & 69.8 \bigstrut[t]\\
			MTL   & V+G+IR & 54.5  & 36    & 92.4  & \underline{72.8} \bigstrut[b]\\
			\hline
			\hline
			\multicolumn{6}{|c|}{\textit{Multiple Visual Feature Aggregation}} \bigstrut\\
			\hline
			HTM  & C3D+V+R & 54.7  & 35.2  & 91.3    & 72.5 \bigstrut[t]\\
			SibNet  & G & 54.2  & 34.8    & 88.2  & 71.7 \\
			POS-CG & IR+M  & 53.9  & 34.9  & 91    & 72.1 \bigstrut[b]\\
			\hline
			\hline
			\multicolumn{6}{|c|}{\textit{VSLAN (ours) with Crossentropy (XE) Loss}} \bigstrut\\
			\hline
			VSLAN\textsubscript{L\textsuperscript{0}} & C3D     & 53.5  & 32.7  & 88.2  & 70.8 \bigstrut[t]\\
			VSLAN\textsubscript{L\textsuperscript{1}} & R & 54.1  & 33.5  & 89.8  & 72.5 \\
			VSLAN\textsubscript{L\textsuperscript{2}} & VGG16 & 48.7  & 31.2  & 84.6  & 68.3 \\
			VSLAN\textsubscript{L\textsuperscript{3}} & FR    & 49.8  & 31.9  & 85.8  & 69.9 \\
			VSLAN\textsubscript{L\textsuperscript{all}} (w/o D. LAN) & R+C3D+V+FR & 55.2  & 35.8  & 94.5  & 73.8 \\
			VSLAN\textsubscript{L\textsuperscript{all}} (w/ D. LAN) & R+C3D+V+FR & 56.4  & 36.5  & 96.8  & 74.3 \bigstrut[b]\\
			\hline
			\hline
			\multicolumn{6}{|c|}{\textit{VSLAN (ours) with Reinforcement Learning (RL) Loss}} \bigstrut\\
			\hline
			VSLAN\textsubscript{L\textsuperscript{all}} (w/o D. LAN) & R+C3D+V+FR & 56.9  & 35.6  & 95.1  & 73.9 \bigstrut[t]\\
			VSLAN\textsubscript{L\textsuperscript{all}} (w/ D. LAN) & R+C3D+V+FR & 57.4  & 36.7  & 97.9  & 75 \bigstrut[b]\\
			\hline
			\hline
			\multicolumn{6}{|c|}{\textit{VSLAN (ours) with and Shared Loss (SL)}} \bigstrut\\
			\hline
			VSLAN\textsubscript{L\textsuperscript{all}} (w/o D. LAN) & R+C3D+V+FR & 57    & 36.1  & 97.3  & 74.2 \bigstrut[t]\\
			VSLAN\textsubscript{L\textsuperscript{all}} (w/ D. LAN) & R+C3D+V+FR & \textbf{57.4} & \textbf{36.9} & \textbf{98.1} & \textbf{75.6} \bigstrut[b]\\
			\hline
		\end{tabular}
	}
	
		\caption{Performance comparison of VSLAN (last $10$ rows) and other state-of-the-art models on MSVD, grouped by segments. Here, BLEU-4 (B@4), METEOR (M), CIDEr (C), and ROUGE\textsubscript{L} (R) are reported as percentage ($\%$). V, G, Y, IR, M, R, FR corresponds to VGG16, GoogleNet, YOLO, Inception ResNet-v2, Optical Flow, ResNext-101, and Faster-RCNN.}
	\label{msvdcomp}
	
\end{table}

\textbf{Training Parameters:} For LAN, we set the unified attention size $z=1024$ and latent dimension $z' = 512$. For bilinear pooling in Equation \ref{eqn:1} and \ref{eqn:2}, we use ELU as activation function, where ReLU for local feature attention at Equation \ref{eqn:3}. To squeeze information, we set $x=256$, $y=512$ for FAN. We transform feature dimension ${L^m}$ and ${{\hat G}_m}$ into similar key $l$, query $q$, value $v$  dimension, $1024$. For VaPen Encoding, the latent distribution ${\delta _t}$ dimension is set to $64$ The decoder LSTM hidden dimension is set to $1024$. We use Adam with $0.0001$ learning rate. For training, we use $64$ as batch size with early-stopping epochs. Both encoder and decoder parameters are uniformly distributed with gradient clip of size $10.0$. Considering the learning uncertainty, before RL training, we pretrain the model with cross-entropy loss up to $10$ epoch for faster convergence. Before training the decoder, we early train VaPEn with $50$ epochs. This warm-up training is done to ease the complete model's training process by initializing the time-consuming latent distribution into a certain state. For shared learning, we set $\eta = 0.3$  empirically. The model is implemented on PyTorch \footnote{https://pytorch.org} with NVIDIA Tesla V100 GPU. VSLAN (full) model consists of a total of $11.6$ million parameters, and one forward pass takes around $10.5$ seconds (averaged over 500 test iterations). In the \textbf{supplementary material}, we plot and analyze the convergence scenario of cross-entropy and reinforcement learning, followed by a discussion on the advantage of shared learning strategy for different $\eta$.

\subsection{Experimental Setup} 

To compare our approach with existing works, we use four standard metrics, BLEU, METEOR, CIDEr, and ROUGE\textsubscript{L} available in MS-COCO toolkit \footnote{https://github.com/tylin/coco-caption}. For both datasets, the best-performing training model on the validation set is applied to the test set for comparison. During testing, we sample captions using beam search with size $5$. For diversity evaluation, we sample 10 POS sequences using VaPEn. Similarly, 10 samples are extracted using DBS.

\subsection{Performance Comparison}

\noindent$(1)$ \textbf{Encoder-Decoder with Attention:} DenseLSTM \cite{mmdenseatt}, SAM \cite{mmsalient}, X-LAN \cite{xlan}, and VRE \cite{mmtwice}. $(2)$ \textbf{Explicit Attribute Guidance:} GRU-EVE \cite{aafaq2019spatio}, HMM \cite{mmhmm}, and C-R \cite{hou2020joint}. $(3)$ \textbf{Encoder-Decoder with Reconstruction:}  RecNet\textsubscript{local} \cite{wang2018reconstruction} and MTL \cite{Pasunuru2017MultiTaskVC}. $(4)$ \textbf{Multiple Visual Feature Aggregation:} HTM \cite{mmhtm}, XlanV \cite{mmxlan}, SibNet \cite{mmsibnet}, and POS-CG \cite{wang2019controllable}. $(5)$ \textbf{Multi Modal Feature Fusion:} M\&M TGMM \cite{chen2017video}, Attention Fusion \cite{hori2017attention}, V-ShaWei-GA \cite{hao2017integrating}, SMCG \cite{mmsibnet}, and MGSA\textsubscript{+audio} \cite{chen2019motion}.

\begin{table}[t]
	
	\centering
	\small
	\scalebox{0.82}{
		\begin{tabular}{|l|c|c|c|c|c|}
			\hline
			\multicolumn{1}{|c|}{\textbf{Model}} & \textbf{Backbone} & \textbf{B@4} & \textbf{M} & \textbf{C} & \textbf{R} \bigstrut\\
			\hline
			\hline
			\multicolumn{6}{|c|}{\textit{Encoder-Decoder with Attention}} \bigstrut\\
			\hline
			\multicolumn{1}{|c|}{DenseLSTM} & C3D+V & 38.1  & 26.6  & 42.8  & - \bigstrut[t]\\
			\multicolumn{1}{|c|}{VRE} & R & 43.2  & 28.0  & 48.3  & 62.0 \bigstrut[t] \bigstrut[b]\\
			\hline
			\hline
			\multicolumn{6}{|c|}{\textit{Explicit Attribute Guidance		}} \bigstrut\\
			\hline
			\multicolumn{1}{|c|}{GRU-EVE} & IR+C3D+Y & 38.3  & 28.4  & 48.1  & 60.7 \bigstrut[t]\\
			\multicolumn{1}{|c|}{HMM} & C3D+IR & 39.9  & 28.3  & 40.9  & - 
			\bigstrut[b]\\
			\hline
			\hline
			\multicolumn{6}{|c|}{\textit{Encoder-Decoder with Reconstruction}} \bigstrut\\
			\hline
			\multicolumn{1}{|c|}{RecNet\textsubscript{local}} & V+IR  & 39.1  & 26.6  & 42.7  & 59.3 \bigstrut[t]\\
			\multicolumn{1}{|c|}{MTL} & V+G+IR & 40.8  & 28.8  & 47.1  & 60.2 \bigstrut[b]\\
			\hline
			\hline
			\multicolumn{6}{|c|}{\textit{Multiple Visual Feature Aggregation}} \bigstrut\\
			\hline
			\multicolumn{1}{|c|}{SibNet} & G & 40.9  & 27.5  & 47.5  & 60.2 \bigstrut[t]\\
			\multicolumn{1}{|c|}{XlanV\textsubscript{RL}} & IR+I3D & 41.2  & 28.6  & 54.2  & 61.5 \\
			\multicolumn{1}{|c|}{POS-CG} & IR+M  & 41.3  & 28.7  & \underline{53.4} & 62.1 \bigstrut[b]\\
			\hline
			\hline
			\multicolumn{6}{|c|}{\textit{Multi Modal Feature Fusion}} \bigstrut\\
			\hline
			\multicolumn{1}{|c|}{M\&M TGMM} & IR+C3D+A & 44.3  & 29.4  & 49.3  & \underline{62.1} \bigstrut[t]\\
			\multicolumn{1}{|c|}{Attention Fusion} & V+C3D+A & 39.7  & 25.5  & 40    & - \\
			\multicolumn{1}{|c|}{V-ShaWei-GA} & C3D+A & \textbf{47.9} & \underline{30.9} & -     & - \\
			\multicolumn{1}{|c|}{MGSA\textsubscript{+audio}} & IR+C3D+A & 45.4  & 28.6  & 50.1  & - \bigstrut[b]\\
			\hline
			\hline
			\multicolumn{6}{|c|}{\textit{VSLAN with Reinforcement Learning and Shared Loss}} \bigstrut\\
			\hline
			VSLAN\textsubscript{L\textsuperscript{1}} & R & 39.8  & 26.5  & 43.7  & 58.1 \bigstrut[t]\\
			VSLAN\textsubscript{L\textsuperscript{all}} (CONCAT) & R+C3D+V+FR & 41.3  & 27.3  & 47.5  & 59.4 \bigstrut[t]\\
			VSLAN\textsubscript{L\textsuperscript{all}} (w/o D. LAN) & R+C3D+V+FR & 45.1  & 31.2  & 54.6  & 60.5 \bigstrut[t]\\
			VSLAN\textsubscript{L\textsuperscript{all}} (w/ D. LAN) & R+C3D+V+FR & \underline{46.5} & \textbf{32.8} & \textbf{55.8} & \textbf{62.4} \bigstrut[b]\\
			\hline
		\end{tabular}
	}
	
	\caption{Performance comparison of VSLAN (last $4$ rows) and other state-of-the-art models on MSR-VTT dataset. The scores are reported as percentage ($\%$). A denotes Audio features.}
	\label{msrvttcomp}
	
\end{table}
\noindent\textbf{MSVD Dataset:} 
We outline the comparison of state-of-the-art methods with VSLAN in Table \ref{msvdcomp}. The last $10$ rows depict different variations of VSLAN. $L\textsuperscript{all}$ stands for the complete VSLAN model utilizing $4$ feature sets. At decoding, we experiment by excluding (w/o D. LAN) and including (w/ D. LAN) the LAN. The model is trained with the losses: Cross-Entropy, Reinforcement, and Shared Loss. The captions of the results are an average score of $10$ VaPEn sample.

The simplest version of our model, VSLAN\textsubscript{L\textsuperscript{0}} achieves an $88.2$ CIDEr score and outperforms approaches based on straightforward attention with encoder and decoder. VSLAN\textsubscript{L\textsuperscript{all}} with XE loss reaches the highest baseline, C-R, with respect to all metrics. Here, we note that the C-R model explicitly guides the model based on external representation, i.e., visual relationship detector using language, where VSLAN solely relies on the video information as input. Similarly, GRU-EVE and HMM guide the model based on attribute information from object and action class labels, where we encourage VSLAN to learn from training captions directly. VSLAN\textsubscript{L\textsuperscript{all}} with LAN decoder, trained with shared loss, gains the best performance compared to approaches to date. Our closest competitor, POS-CG, utilizes training captions to extract POS information and then employ multiple visual features to predict POS, later used to generate the captions. We rather exploit only the visual information available and leave the pos-guided caption generation end-to-end, which gains $+7.8\%$ CIDEr score over POS-CG. Also, robustness of our VSLAN Encoder is identified by using a regular decoder on the average global feature, $\hat G$ without attention (w/o D. LAN), which outperforms attention and reconstruction methods by a wide margin.

\begin{table}[t]
	
	\centering
	\small
	\scalebox{0.7}{ 
		\begin{tabular}{|c|c|c|c|c|c|c|}
			\cline{2-7}\multicolumn{1}{r|}{} & \multicolumn{3}{c|}{MSVD} & \multicolumn{3}{c|}{MSR-VTT} \bigstrut\\
			\hline
			\textbf{Model} & \textbf{mBleu-4} & \textbf{Div-1} & \textbf{Div-2} & \textbf{mBleu-4} & \textbf{Div-1} & \textbf{Div-2} \bigstrut\\
			\hline
			\hline
			RecNet & 0.86  & 0.22  & 0.23  & 0.84  & 0.2   & 0.21 \bigstrut[t]\\
			POS-CG & 0.79  & 0.24  & 0.25  & 0.76  & 0.22  & 0.23 \\
			X-LAN & 0.83  & 0.22  & 0.25  & 0.8   & 0.24  & 0.28 \\
			SMCG+AllRec & 0.74  & 0.28  & 0.31  & 0.61   & 0.27  & 0.29 \\
			VSLAN\textsubscript{no VaPEn} (w/o D. LAN) & 0.82  & 0.21  & 0.23  & 0.81  & 0.23  & 0.26 \\
			VSLAN\textsubscript{no VaPEn} (w D. LAN) & 0.77  & 0.26  & 0.29  & 0.75  & 0.26  & 0.29 \\
			VSLAN\textsubscript{full}  (w/o DBS) & 0.64  & 0.31  & \textbf{0.36} & 0.63  & 0.28  & 0.31 \\
			VSLAN\textsubscript{full}  (w DBS) & \textbf{0.62} & \textbf{0.32} & 0.36  & \textbf{0.58} & \textbf{0.3} & \textbf{0.33} \bigstrut[b]\\
			\hline
			\hline
		\end{tabular}

	}
	
	\caption{Comparison of related methods on diversity metrics.}
	\label{diversity}
	
\end{table}

\textbf{MSR-VTT Dataset:}

This dataset is relatively complex with context and audio information. As mentioned earlier, we ignored modalities other than exploiting only visual data of MSR-VTT. For comparison, we train the best models on the MSVD dataset, mentioned in the last $4$ rows of Table \ref{msrvttcomp}. GRU-EVE and HMM, even guide training with attributes, exploit only visual features, are outperformed by $7\%$ using VSLAN with basic decoder (w/o D. LAN). Similarly, VSLAN with a basic decoder excels in reconstruction models, which uses a comparatively complex decoder. Similar performance is achieved against the multiple visual feature fusion models. However, recent state-of-the-art works are based on multi-modal, i.e., audio, category information with visual data. To verify our model robustness, we compare with state-of-the-art multi-modal models. We notice that our basic decoder ranks almost similar to the closest competitor, V-ShaWei-GA. However, the full model (w/ D. LAN) slightly improves the basic decoder. Our best performing metric is CIDEr, which outperforms POS-CG and MGSA\textsubscript{+audio} by $4.5\%$ and $11.4\%$, respectively. We argue that rich audio information of the MSR-VTT leveraged better BLEU on V-ShaWei-GA. It is worth noting that our result does not necessarily require any sophisticated pre-trained CNN models because of our inherent architecture's feature distillation process. We note that VSLAN trained with R has also achieved competitive performance with a concatenated feature only (CONCAT, w/o. FAN). Thus the steady improvement can be directly linked to our VSLAN architecture, which results in fine-grained representation. Extended VSLAN with category and audio info is expected to gain better results, which is out of our research scope.

\begin{figure*}[t]
	\centering
	\includegraphics[width=0.75\textwidth]{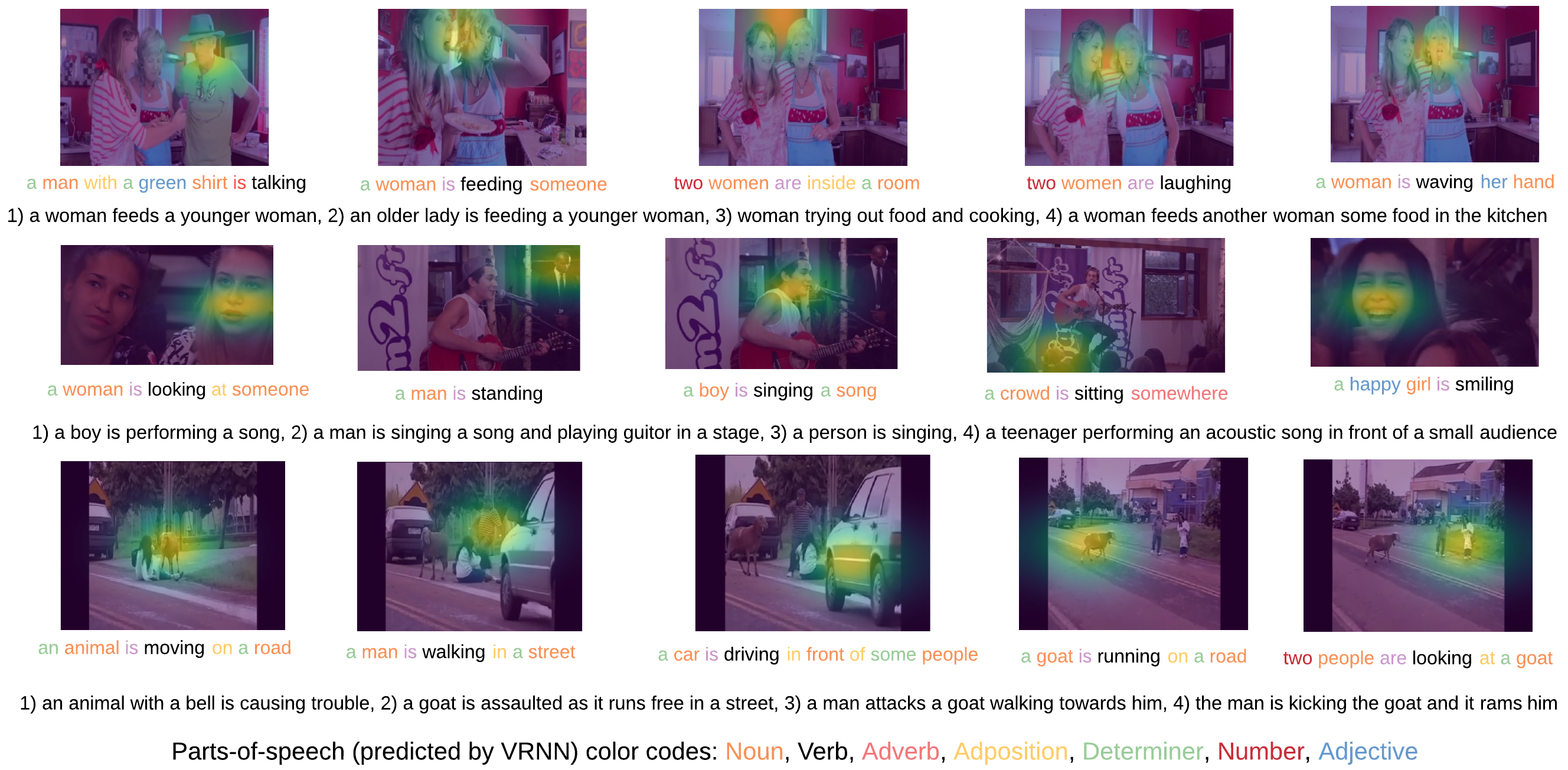}
	
	\caption{A detailed visual illustration of the diverse captions generated by our architecture. Each row of images represent sequential frames of a random test video. The heatmap depicts spatio-temporal attention distribution while generating each sentence (showed below the images). The color codes of each caption represents POS by VaPEn, in the bottom. For each, four reference ground truth are presented.}
	\label{fig:qual_analysis_detailed}
\end{figure*}

\textbf{Diversity Analysis:} In Table \ref{msvdcomp} and \ref{msrvttcomp}, we see VSLAN's overall performance. However, these evaluations are limited to the efficacy of LAN and FAN block. Because, for these evaluations, we sample top-1 caption. So, the full potential of VaPEn is ignored. To this end, we evaluate the diversity of VSLAN's captions in Table \ref{diversity}. We use two protocols: mBleu-4 (lower is better), Div-n (higher is better). m-Bleu-4 performs a cross-match of 4-gram occurrences on the generated captions from one clip. These metrics ensure that the sentences are significantly different from each other. Div-n indicates a unique n-gram ratio between the generated captions. For more details regarding the metrics, we refer readers to \cite{jyoti}. We compare four versions of our model with recent algorithms, RecNet, POS-CG, and SMCG. Additionally, we use the pre-trained model of XlanV\textsubscript{RL} on our datasets to compare. VSLAN\textsubscript{no VaPEn} is VSLAN without VaPEn, where we map $\tilde{G}$ to $\bar{G}$ using a linear layer. We can analyze that VSLAN\textsubscript{no VaPEn} with decoder LAN achieves marginal performance compared to others. So, adding a decoder does not guarantee better diversity. However, when the VaPEn is included, the performance jumps and exceeds the compared methods. This phenomenon is due to the probabilistic property of the VaPEn encoder, which encourages the decoder to generate diverse captions. Moreover, we can see that DBS does not have a larger effect compared to VaPEn. Finally, we can see that the captions are syntactically correct (based on CIDEr scores) and diversified (based on the diversity evaluation).

\begin{figure}[h]
	\centering
	\includegraphics[width=0.9\columnwidth]{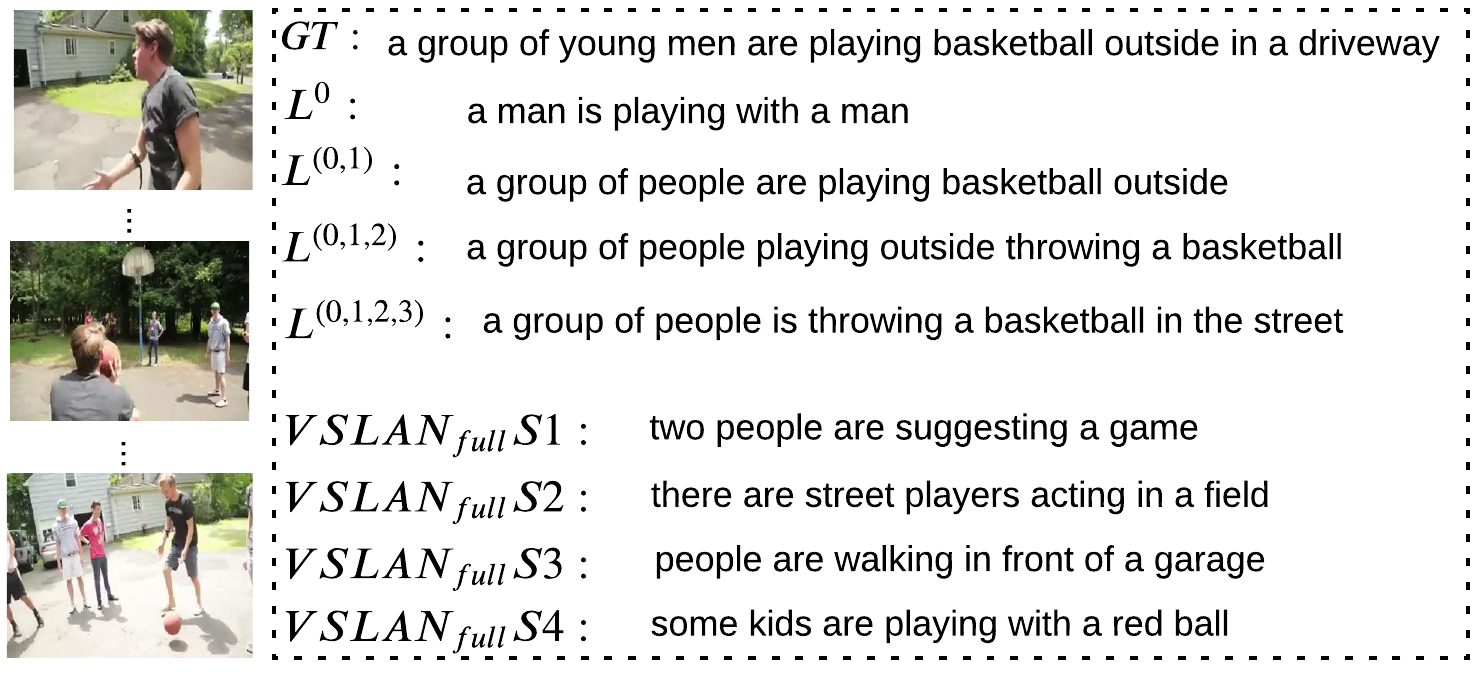}

	\caption{Qualitative evaluation on a random test video. $GT$ is the ground truth, $L^m$ are the indices of used feature sets.}
	\label{fig:qablock}
\end{figure}

\textbf{Qualitative Analysis:} Figure \ref{fig:qablock} shows a ground truth $GT$ sentence and predicted captions for a random test video. The respective feature sets, $L^m$, are propagated through stacking the layers. We notice that $L^0$ misses basic object properties, where $L^1$ with new block corrects sentence structure. With $L^{(0,1,2)}$, additional subject and action features are attended, which results in a new verb, ``throwing." Finally, the ``street" attribute is captured by $L^{(0,1,2,3)}$, where Faster R-CNN was introduced. The result evolution depicts the robustness of feature interaction while passing novel set using FAN block. Some samples from VSLAN are shown in the last 4 rows. Figure \ref{fig:qual_analysis_detailed} organizes our complete model performance into one page. For each row of frames (belong to a single video), it can be seen that each sample of VaPEn influences the caption decoder to attend different spatio-temporal regions behind generating each diverse caption. For example, in row 2, ``a man is standing" and ``a boy is singing a song" belong to the nearby frame, whereas two attention locations influenced different meaningful yet relevant captions. Similarly, in the last row, we can see a failure case, where our method could not attend to appropriate locations due to lack of training. For each row sequence, we showed four ground truth captions as a reference. We have further evaluated the consistency of VSLAN based on two ablation studies. First, if the performance remains consistent upon shuffle of the feature orders. Second, if there is any performance gain if we use the existing model trained on video captioning dataset, followed by a cross-dataset experiment. We refer readers to the \textbf{supplementary material}.

\section{Conclusion}

This paper proposes VSLAN for generating diverse captions by exploiting explicit feature interaction and aggregating multiple feature streams in a discount fashion, followed by diverse POS prediction. Previous captioning models aggregated features from multiple extractors by either weighted concatenation or passing through linear layers to produce new representations. Also, they ignored diversity in caption generation. Whereas, VSLAN's learned attributes overcome the previous approaches' limitation on generating diverse captions by attending temporal states end-to-end, which achieved state-of-the-art performance.

{\small
	\bibliographystyle{ieee_fullname}
	\bibliography{egbib}
}

\end{document}